\documentclass[10pt,twocolumn,letterpaper]{article}

\usepackage[pagenumbers]{wacv} 

\usepackage{graphicx}
\usepackage{amsmath}
\usepackage{amssymb}
\usepackage{booktabs}
\usepackage{pifont}

\usepackage{graphicx}
\usepackage{amsmath}
\usepackage{amssymb}
\usepackage{booktabs}

\usepackage{graphicx}

\newcommand{\temp}{-1em}

\newcommand{\offset}{\theta_\text{offset}}
\newcommand{\offsetqg}{\theta^z_\text{offset}}
\newcommand{\encoderswin}{\mathcal{E}^\theta_\text{SWIN}}
\newcommand{\encodercf}{\mathcal{E}^\theta_\text{ChartEncoder}}
\newcommand{\Model}{\textsc{ChartFormer}}
\newcommand{\Modelvqa}{\textsc{QDChart}}

\newcommand{\cmark}{\ding{51}}
\newcommand{\xmark}{\ding{55}}

\newcommand{\bsb}{\boldsymbol}
\newcommand{\real}{\mathbb{R}}

\usepackage{booktabs}

\usepackage[pagebackref,breaklinks,colorlinks]{hyperref}

\usepackage[capitalize]{cleveref}
\crefname{section}{Sec.}{Secs.}
\Crefname{section}{Section}{Sections}
\Crefname{table}{Table}{Tables}
\crefname{table}{Tab.}{Tabs.}

\def\wacvPaperID{1615} 
\def\confName{WACV}
\def\confYear{2025}

\begin{document}

\title{Advancing Chart Question Answering with\\Robust Chart Component Recognition}

\author{Hanwen Zheng\\
Virginia Tech\\
{\tt\small zoez@vt.edu}
\and
Sijia Wang\\
Virginia Tech\\
{\tt\small sijiawang@vt.edu}
\and
Chris Thomas\\
Virginia Tech\\
{\tt\small christhomas@vt.edu}
\and
Lifu Huang\\
Virginia Tech\\
{\tt\small lifuh@vt.edu}
}
\maketitle

\begin{abstract}
Chart comprehension presents significant challenges for machine learning models due to the diverse and intricate shapes of charts. Existing multimodal methods often overlook these visual features or fail to integrate them effectively for chart question answering (ChartQA). To address this, we introduce \Model{}, a unified framework that enhances chart component recognition by accurately identifying and classifying components such as bars, lines, pies, titles, legends, and axes. Additionally, we propose a novel Question-guided Deformable Co-Attention (QDCAt) mechanism, which fuses chart features encoded by \Model{} with the given question, leveraging the question's guidance to ground the correct answer. Extensive experiments demonstrate that the proposed approaches significantly outperform baseline models in chart component recognition and ChartQA tasks, achieving improvements of 3.2\% in mAP and 15.4\% in accuracy, respectively. These results underscore the robustness of our solution for detailed visual data interpretation across various applications.

\end{abstract}

\section{Introduction}

\begin{figure}[t!]
    \centering
    \includegraphics[width=0.45\textwidth]{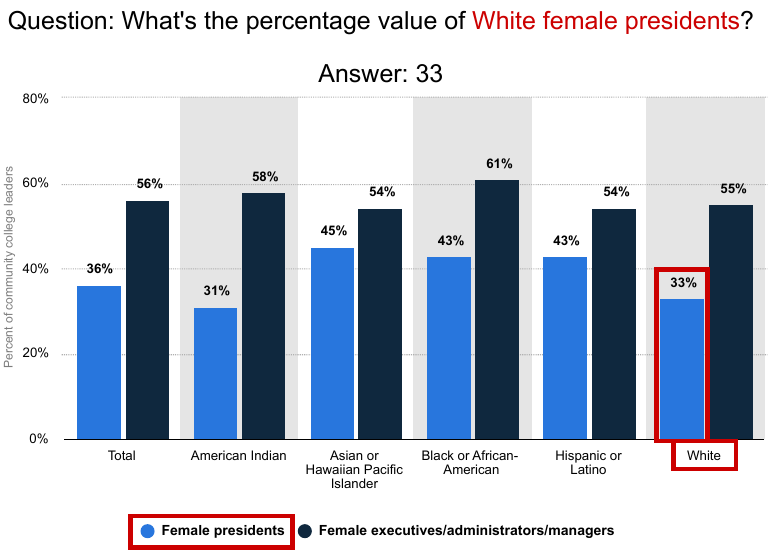}
    \caption{An example of Chart Question Answering.}
    \label{fig:chartqa_moti}
\end{figure}

Comprehending charts and correctly answering chart-related questions~\cite{lee_pix2struct_2022, masry_chartqa_2022, cheng_chartreader_2023} is essential in today's data-driven world. Charts are powerful tools to distill complex data into visual formats, making it easier to identify trends, patterns, and insights at a glance. Despite the significant progress that researchers have made on Visual Question Answering~\cite{agrawal_vqa_2016,ding_mukea_2022,yang_stacked_2016,shen2024multimodal,xu2024vision,xu2022multiinstruct,qi2023art,reddy2022mumuqa}, Chart Question Answering (i.e., ChartQA) is particularly challenging as it requires seamless and fine-grained interpretation and analysis of both textual and visual elements in the charts to answer natural language questions~\cite{masry_chartqa_2022, methani_plotqa_2020,kim_ocr-free_2022,cheng_chartreader_2023,lee_pix2struct_2022,liu_matcha_2022,masry_unichart_2023,zhou_enhanced_2023}. Consider the example shown in Figure~\ref{fig:chartqa_moti}. To correctly answer the question based on the given chart, models need to accurately locate textual elements such as ``\textit{White}'' in the category axis and ``\textit{Female presidents}'' in the legend. In addition, they also need to identify the visual elements such as the blue bar for ``\textit{White}'' category and correctly link it to the corresponding text label, i.e., ``\textit{33\%}''.

Though many efforts have been made on chart component recognition to correctly identify key components such as chart types (e.g., \textit{pie}, \textit{bar}, and \textit{line} charts), visual elements in each plot (e.g., \textit{connected lines}, \textit{colors}), and textual elements (e.g., \textit{legends} and \textit{value axes}), they usually require 
a pipeline approach that first detects key points of lines or boxes and then classifies and groups the detected regions. 
Such methods struggle to comprehend complex graphics, such as stacked boxes, overlapped labels, and crossed lines. 
For ChartQA, it is essential to incorporate rich and accurate chart information and ensure proper attention is given to the relevant chart regions based on the specific questions asked.
Existing approaches that leverage OCR tools or extracted tables often introduce noise and confusion due to inaccuracies. Additionally, existing models typically fuse the visual information with the question embedding at a later stage, relinquishing the ability to guide attention to the relevant parts of the graphics via the question.

To tackle these challenges, we propose a novel framework named \Modelvqa{}, which integrates an innovative unified solution, \Model{}, to recognize all the various chart components from diverse types of charts. \Model{} leverages  
deformable attention \cite{xia_dat_2023} to effectively capture the visuals of chart graphics, and incorporates a mask attention mechanism \cite{cheng_masked-attention_2022} with learnable query features to cover diverse chart components.
To address ChartQA, we propose a novel Question-guided Deformable Co-Attention (QDCAt) mechanism that leverages the rich chart features encoded by \Model{}. This mechanism fuses question information with chart features through a Question-guided Offset Network (QON) and integrates visual and chart-related features using a deformable co-attention module. The resulting question-guided features are then passed to a text decoder to generate the answer to the given question.

We first evaluate the effectiveness of \Model{} on chart component recognition, on the public benchmark dataset ExcelChart400K \cite{luo_chartocr_2021}. \Model{} significantly outperforms existing strong baselines, such as DAT~\cite{xia_dat_2023} Mask2former~\cite{cheng_masked-attention_2022}, by 11.4\% and 3.2\% in mAP, respectively. It especially shows superior performance on stacked bars, overlapped or fluctuating lines, and narrow pie slices.
We further evaluate \Modelvqa{} on ChartQA~\cite{masry_chartqa_2022}, a large dataset that includes human-written and machine-generated questions. \Modelvqa{} outperforms the previous strong baseline Pix2Struct~\cite{lee_pix2struct_2022} by $1.2\%$ accuracy. Our model excels in handling visually related questions, particularly those involving color disambiguation or size comparison of chart components. Additionally, it demonstrates superior ability in effectively grounding key chart information.

The contribution of this work can be summarized as follows: 
\begin{itemize}
\item \Model{}, a state-of-the-art end-to-end model for unified chart component recognition, provides accurate identification for diverse class objects with distinct visual structures and rich chart semantics.
\item \Modelvqa{}, a multimodal application leveraging \Model{}'s rich chart semantics with a novel Question-guided Deformable Co-Attention (QDCAt) fusion layer for ChartQA, enables the model to focus on components related to the question.
\item \Model{} exceeds the baseline model by 3.2\% in mAP on chart component recognition, while \Modelvqa{} surpasses the baseline model by 15.4\% in accuracy on ChartQA.
\item An automatically annotated unified chart component recognition instance segmentation dataset featuring several prominent chart types.
\end{itemize}

\section{Related Work}

\paragraph{Chart Component Recognition}
Existing approaches for chart element recognition can be divided into three broad categories: detection via bounding boxes \cite{ma_towards_2021,liu2019dataextractionchartssingle,choi_visualizing_2019}, key point detection and grouping \cite{luo_chartocr_2021, xue_chartdetr_2023,cheng_chartreader_2023,ma_towards_2021}, and line graph detection \cite{lal_lineformer_2023,p_lineex_2023}. Among these, {ChartOCR} \cite{luo_chartocr_2021} uses an hourglass network to identify key points and a rule-based approach to associate them with chart elements.  
{Lenovo} \cite{ma_towards_2021} trains separate detectors for points and bars, followed by a deep neural network to measure feature similarities for data conversion.
{ChartDETR} \cite{xue_chartdetr_2023} employs a transformer-encoder-decoder model to detect and categorize key point groups within a unified framework.
{Lineformer} \cite{lal_lineformer_2023} focuses on line charts, treating line detection as instance segmentation using the mask2former model~\cite{cheng_masked-attention_2022}. Compared to previous studies, 
our proposed \Model{} instead employs a comprehensive end-to-end instance segmentation framework to enhance chart component recognition.

\paragraph{Object Detection and Instance Segmentation}
Object detection and instance segmentation \cite{10155181, 10.1115/IMECE2021-69975, 10.1007/978-3-031-25066-8_41} are core tasks in computer vision and have been widely explored by using Convolutional Neural Networks~\cite{schmidhuber_deep_2015,he_mask_2018,he_deep_2015,ren_faster_2016,wang_solov2_2020}. 
Recently, many new approaches have been developed based on Transformer~\cite{9564709,Zhang_2023_CVPR}, inspired by its success in the field of Natural Language Processing. Among them, ViT~\cite{dosovitskiy_image_2021} processes images as non-overlapping patch sequences, using global attention to capture long-range dependencies. Swin Transformer~\cite{liu_swin_2021}, on the other hand, employs partitioned window attention to focus on specific regions within images. Enhanced attention mechanisms like the Deformable Convolutional Network (DCN)~\cite{dai_deformable_2017} and Masked attention Mask Transformer (Mask2Former)~\cite{cheng_masked-attention_2022} are further introduced to enhance these approaches. DAT (Deformable Attention Transformer)~\cite{xia_dat_2023} improves DCN's capabilities by learning deformed key points through feature sampling and updating positional embeddings with relative position bias. Our work leverages the strengths of instance segmentation frameworks on small and varied objects to improve chart understanding.

\begin{figure*}[t]
    \centering
    \includegraphics[width=1.0\textwidth]{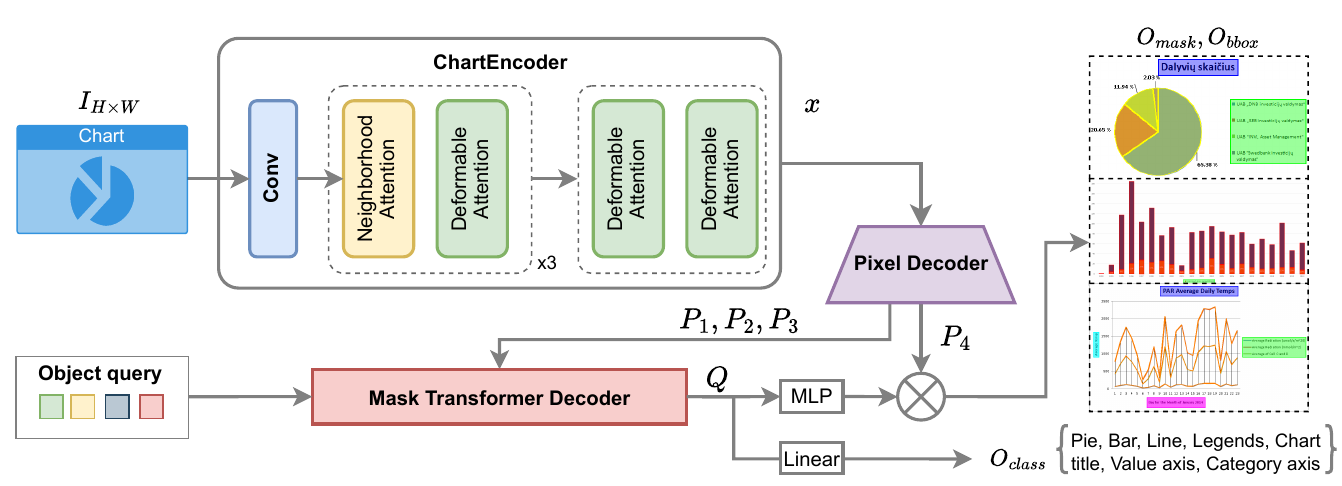}
    \caption{Model architecture of \Model{} for chart component recognition}
    \label{fig:chart_model}
\end{figure*}

\paragraph{Chart Question Answering}
Recent advancements in language and multimodal models have significantly enhanced their ability to tackle the complex reasoning required for ChartQA tasks. {Donut}~\cite{kim_ocr-free_2022} is a vision-encoder-text-decoder model that leverages Swin Transformer~\cite{liu_swin_2021} and MBart~\cite{liu_multilingual_2020} to answer questions with a visual context. 
{VL-T5} \cite{cho_unifying_2021} extends \cite{raffel_exploring_2020} by incorporating visual features from chart images, 
while {VisionTaPas} \cite{masry_chartqa_2022} extends TaPas~\cite{herzig_tapas_2020} by integrating a vision transformer encoder to process chart images. {ChartT5} \cite{zhou_enhanced_2023} improves chart understanding by leveraging a visual and language pre-training framework on chart images and predicted table pairs. 
{ChartReader}~\cite{cheng_chartreader_2023} leverages a transformer-based model to detect chart components and {{Pix2Struct}~\cite{lee_pix2struct_2022} adopts an image-encoder-text-decoder architecture and introduces a screenshot parsing pre-training objective based on the HTML source of web pages, aiming to enhance the model's layout and language understanding capabilities.} MatCha~\cite{liu_matcha_2022} utilizes two pre-training tasks, chart derendering and math reasoning, to enhance visual language understanding. Similarly, UniChart \cite{masry_unichart_2023} enhances Donut's capabilities by pre-training on four chart-related tasks: data table generation, numerical \& visual reasoning, open-ended question answering, and chart summarization.
Compared to these existing efforts, \Modelvqa{} employs a novel unified solution for detecting all the chart components and further enhances the visual reasoning capability by attending to the components that are particularly related to questions.

\section{Method}
In this section, we first introduce \Model{}, the first unified and end-to-end solution for recognizing components from diverse types of charts with distinct visual structures (Section \ref{sec:chartformer}), and then illustrate \Modelvqa{} which integrates the chart components identified by \Model{} and selectively incorporates them to answer the target question (Section~\ref{sec:modelvqa}).

\subsection{\Model}
\label{sec:chartformer}

As shown in Figure~\ref{fig:chart_model}, given an input chart image $\bsb{I}$ with dimensions \(H \times W\), \textbf{chart component recognition} aims to identify and classify various components within the chart, including both visual elements such as \textit{lines}, \textit{bars}, and \textit{pie slices}, and textual elements such as \textit{value axes}, \textit{category axes}, \textit{legends}, and \textit{chart titles}. 
Let $C$ be the set of all chart component types, e.g., $C = \{\text{"Bar"}, \text{"Line"}, \text{"Pie"}, \ldots\}$. 
Let \(M_c\) represent the set of instance segmentation regions corresponding to a particular chart component type \(c \in C\), and $M = \{ M_c \}_{c \in C}$ be the set of all annotations for the image $\bsb{I}$.

\Model{} is designed to learn to accurately identify and classify each chart component in chart images, and consists of three main modules: a vision encoder, a pixel decoder, and a mask transformer decoder, as shown in Figure~\ref{fig:chart_model}. Formally, the input chart image $\bsb{I} \in \real^{H \times W}$ first undergoes processing by a CNN layer to generate an initial feature map $\bsb{E} \in \real^{K \times H/4 \times W/4}$ with a channel size $K = 64$ in our experimental setting. This feature map $\bsb{E}$ is fed into the \textbf{vision encoder} for further processing and feature learning. The vision encoder features multiple blocks of neighborhood attention \cite{hassani_neighborhood_2022} and deformable attention \cite{xia_dat_2023}. 
Neighborhood attention expands each pixel’s attention span to its nearest neighborhood 
\begin{equation}
    \bsb{\chi} = \text{NeigAttention}(\bsb{E})\ .
\end{equation}
Chart images have distinct borders and consecutive visual elements like lines and bars, thus applying deformable attention becomes intuitive to emphasize precise focus on the visual connections. In deformable attention, uniformly spaced reference points $p$ are offset by $\Delta p$, obtained via an offset network $\offset$ applied to the query vectors $\bsb{q}$.
Features are then computed using bilinear sampling $\phi$ at the deformed points
\begin{equation}\label{eq:dsampling}
    \Delta p = \offset(\bsb{q}), \quad \bsb{\tilde{\chi}} = \phi(\bsb{\chi}; p + \Delta p )\ .
\end{equation}

\begin{figure*}[ht]
    \centering
    \includegraphics[width=\textwidth]{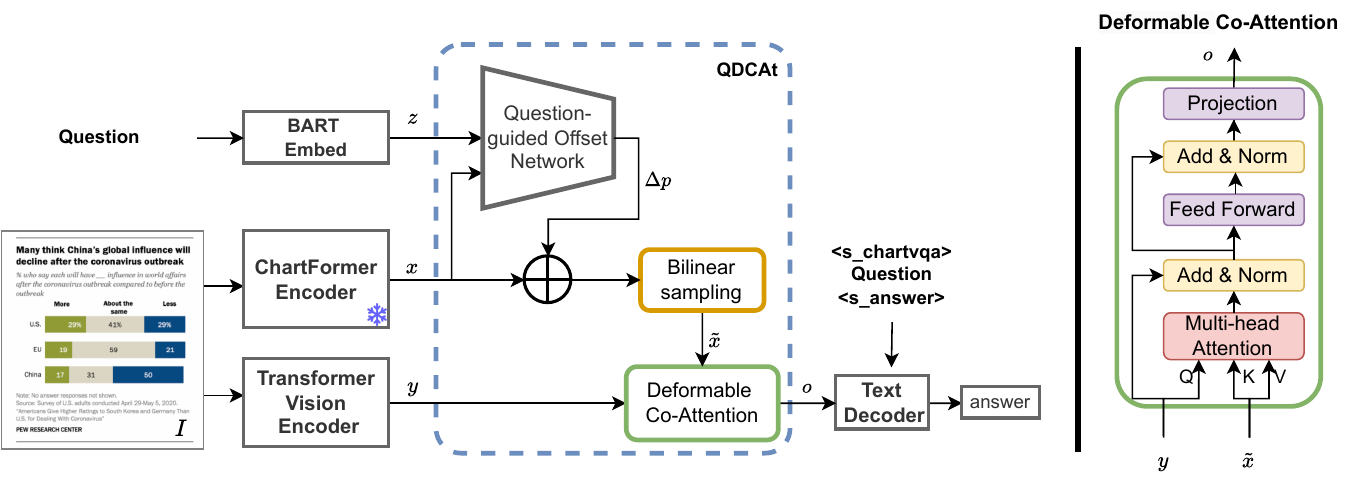}
    \caption{The \Modelvqa{} model structure}
    \label{fig:chartqa_model}
\end{figure*}

Then the deformation attention is computed as
\begin{align}
    \bsb{q} = \bsb{\chi} W_q\ , \quad \bsb{k} &= \bsb{\tilde{\chi}} W_k\ , \quad \bsb{v} = \bsb{\tilde{\chi}} W_v\ ,\\
    \text{DefoAttention}(\bsb{\chi}) &= \text{softmax} \left( \bsb{q} \bsb{k}^\top \!\!/\! \sqrt{d} \right) \bsb{v}\ ,\label{eq:dattention}
\end{align}
where key $\bsb{k}$ and value $\bsb{v}$ vectors are projected from sampled features $\bsb{\tilde{\chi}}$, and $d$ is the dimension of the attention head.

These specialized attention mechanisms allow \Model{} to effectively capture local and global contextual information, thereby enhancing the ability to extract meaningful features from chart images. 
Together, the CNN layer and the vision encoder extract features $\bsb{x} \in \real^{8K \times H/32 \times W/32}$ from the input image $\bsb{I}$
\begin{align}
    \bsb{x} = \encodercf(\bsb{I})\ .
\end{align}

The \textbf{pixel decoder} then upsamples the extracted features $\bsb{x}$ and outputs a 2D per-pixel embedding $\bsb{P} \in \real^{C_E \times HW}$, where $C_E = 256$ is the number of channels 
\begin{equation}
    \bsb{P} = \bsb{P}_4\ , \hspace*{0.5em}
    \bsb{P}_{i} = \mathcal{D}^i_\text{pixel}(\bsb{P}_{i-1})\ , \hspace*{0.5em}
    \bsb{P}_0 = \bsb{x}\ .
\end{equation}

Subsequently, a \textbf{mask transformer decoder} 
combines object queries $\bsb{u} \in \real^{C_Q \times N}$ and pixel decoder features to compute embeddings $\bsb{Q} \in \real^{C_Q \times N}$, where $C_Q = 256$ denotes the number of channels and $N$ denotes the number of object queries
\begin{equation}
    \bsb{Q} = \mathcal{D}_\text{mask}(\bsb{u} ; \bsb{P}_{1:3})\ .
\end{equation}

The embeddings $\bsb{Q}$ are passed through dense layers to predict object classes $\bsb{O}^i_\text{class} \in \{\Delta^L\}_{i=1}^N$ and binary per-pixel mask predictions $\bsb{O}_\text{mask} \in \{0,1\}^{N \times HW}$
\begin{align}
    \bsb{O}^i_\text{class} &= \text{softmax}(\bsb{Q}_i^\top W)\ , \\
    \bsb{O}_\text{mask} &= s \Big( \sigma \big(\bsb{M}^\top \bsb{P}\big) - t \Big)\ ,
\end{align}
where $\sigma$ is the element-wise sigmoid function, $t \in [0, 1]$ is a threshold parameter, $s$ is the step function, and $\bsb{M} = \text{MLP}(\bsb{Q})$. The bounding box $\bsb{O}_{bbox}$ can be directly obtained by drawing the smallest box that encloses the segmentation mask. 
Following \cite{cheng_masked-attention_2022}, we use a combination of focal loss and dice loss for $\bsb{O}_\text{mask}$, and classification loss for $\bsb{O}_\text{class}$.

\subsection{\Modelvqa{}}\label{sec:modelvqa}

Given an input image $\bsb{I}$ and a natural language question $Q$, we further design \Modelvqa{} to provide an answer $A$ by leveraging the chart components detected by \Model{}. As shown in Figure~\ref{fig:chartqa_model}, \Modelvqa{} consists of four main modules: a chart encoder, a vision encoder, a Question-guided Deformable Co-Attention (QDCAt) fusion block that fuses the output of two encoders, and a text decoder.

\paragraph{Chart Encoder}
We include the vision encoder of the \Model{} model ($\encodercf$) as a primary image encoder, utilized to capture explicit chart segment information. 
During the fine-tuning stage on the ChartQA dataset, we freeze the weights $\theta$ of the \Model{} model to reduce training costs.

\paragraph{Vision Encoder}
The Vision Encoder module follows the previous work \cite{kim_ocr-free_2022} and utilizes a {Swin Transformer \cite{liu_swin_2021} architecture to provide complementary visual information, denoted as $\encoderswin(\bsb{I})$}. The parameters are initialized based on the pre-trained model \cite{kim_ocr-free_2022}. We pass the image through the Swin encoder to obtain the feature $\bsb{y}$ 
\begin{equation}
    \bsb{y} = \encoderswin(\bsb{I})\ ,
\end{equation}
where $\theta$ are trainable parameters of the Vision Encoder.

\begin{figure}[t]
    \centering
    \includegraphics[width=0.45\textwidth]{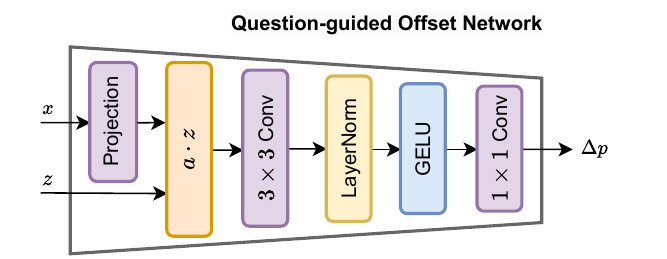}
    \caption{Question-guided offset network}
    \label{fig:qgoff-small}
\end{figure}

\paragraph{\texorpdfstring{\!}{}Question-guided Deformable Co-Attention}

To fuse image features from complimentary encoders, and to incorporate information guided by the question, we propose Question-guided Deformable Co-Attention (\textbf{QDCAt}), which consists of a question-guided offset network and a deformable co-attention block.

To capture complex spatial relationships and patterns in images and associate them with VQA questions, \textbf{question-guided offset network} is proposed. 
The proposed Question-guided offset network $\offsetqg$ is distinct in being conditioned on the question embedding provided in the ChartQA task, as shown in Figure~\ref{fig:chartqa_model}.
We obtain the token embeddings of the question as provided by {the last hidden layer of} MBart \cite{liu_multilingual_2020}

and denote the output as $\bsb{z}$.
The input tensor $\bsb{x}$ is passed through a projection layer\footnote{We use the term ``projection layer'' to refer to a convolution layer with a $1 \times 1$ kernel.} parametrized by the weights $W_a$ to obtain $\bsb{a} = \bsb{x} W_a$.
We take the dot product of $\bsb{z}$ and $\bsb{a}$ to get $\bsb{z} \bsb{a}^\top$, which we pass into a $k \times k$ convolution layer $\text{Conv}$, normalization $\text{Norm}$, $\text{GELU}$ activation, and a projection layer parametrized by the weights $W_{\Delta p^z}$
\begin{align}
     \Delta p^z &= \offsetqg(\bsb{x})\\
     &= \text{GELU}(\text{Norm}(\text{Conv}(\bsb{z} (\bsb{x}W_a)^\top))) W_{\Delta p^z}.\nonumber
\end{align}
The idea behind giving the offset network the combined input of $\bsb{z}$ and $\bsb{a}$ is to make the sampled points 
align with image locations with semantic significance for the question of the particular data sample, as opposed to locations with overall significance. This allows further features learned by the model to represent visual queues with specific relevance to the given question.
A visual representation of the question-guided offset network is provided in Figure~\ref{fig:qgoff-small} as well as Figure~\ref{fig:qgoff} in Appendix~\ref{sec:appendix}. 

The question-guided offsets $\Delta p^z$ are then used to offset per-pixel coordinates $p$, positioning the points used for bilinear sampling
\begin{equation}
    \bsb{\tilde{x}} = \phi(\bsb{x}; p + \Delta p^z)\ .
\end{equation}

Next, we introduce the \textbf{deformable co-attention block}.
This novel attention block integrates and enhances the advantages of co-attention and deformable attention, where
{co-attention} combines features extracted from multiple modalities, and  
deformable attention in formula \eqref{eq:dattention} emphasizes attention computed at semantically relevant features.
The proposed \emph{deformable co-attention} extends this approach by combining features  $\bsb{\tilde{x}}$ and $\bsb{y}$.
\begin{align}
    \bsb{q} = \bsb{y} W_q\ , \quad \bsb{k} = \bsb{\tilde{x}} W_k\ , \quad \bsb{v} = \bsb{\tilde{x}} W_v\ ,\\
    \!\text{DCAttention}(\bsb{x}, \bsb{y}, \bsb{z}) = \text{softmax}\! \left( \bsb{q} \bsb{k}^{\!\top} \!\!/\! \sqrt{d} \right)\! \bsb{v}\ .\!
\end{align}

The output of $\text{DCAttention}(\bsb{x}, \bsb{y}, \bsb{z})$ is passed through residual connections, normalization layers, and a feed-forward network. The Add and LayerNorm operations perform element-wise addition of features and normalization.
Their first occurrence surrounds deformable co-attention as follows:
\begin{equation}
    \!\bsb{b} = \text{LayerNorm}(\bsb{y} \hspace*{-.1em}+\hspace*{-.1em} \text{DCAttention}(\bsb{x}, \hspace*{-.1em}\bsb{y}, \hspace*{-.1em}\bsb{z}))\ .
\end{equation}
The feed-forward network consists of a $3 \times 3$ convolution layer followed by ReLU activation. 
The second occurrence of Add and LayerNorm surrounds the feed-forward network
\begin{equation}
    \bsb{c} = \text{LayerNorm}(\bsb{b} + \text{ReLU} ( \text{CNN}(\bsb{b}) ))\ .
\end{equation}
The output $\bsb{c}$ is lastly projected via $W_o$ to obtain $\bsb{o} = \bsb{c} W_o$, which is the output of the deformable co-attention block as the final image feature.

\paragraph{Text Decoder}
Following the approach of Donut \cite{kim_ocr-free_2022}, we also employ the MBart decoder \cite{liu_multilingual_2020} for answer output generation. Task-specific prompts, for instance, \texttt{<chartvqa>}, \texttt{<s\_answer>} are provided to the decoder as special tokens, and the decoder generates the output by predicting the next token based on the prompted context. 
{The text decoder, denoted as $\mathcal{D}^\theta_\text{text}$, takes the processed feature $o$ by QDCAt and the task-specific prompt $p$ to generate answers $A$}
\begin{equation}
    A = \mathcal{D}^\theta_\text{text}(p; \bsb{o})\ .
\end{equation}

\section{Experimental Setup}
We first evaluate \Model{} on the chart component recognition task and then further assess the effectiveness of \Modelvqa{} on chart question answering.

\subsection{Chart Component Recognition}
\paragraph{Baselines}
We employ four advanced models as strong baselines for chart component recognition and compare them with \Model{}:  
\textbf{Mask R-CNN}, which \cite{he_mask_2018} extends Faster R-CNN by adding a branch for predicting segmentation masks; \textbf{SOLOv2} \cite{wang_solov2_2020} utilized a specialized segmentation branch with decoupled head for better mask feature learning; \textbf{Mask2Former} \cite{cheng_masked-attention_2022}incorporates masked attention mechanisms for unified segmentation tasks; \textbf{DAT} \cite{xia_dat_2023}, characterized by its integration of deformable attention mechanisms and global image features, facilitates the comprehensive analysis of multiple chart components. 

\begin{table*}
\small \centering
{
\begin{tabular}{l|r|r|r|r|r|r}
\toprule
\textbf{Split} & \multicolumn{2}{c|}{\textbf{Train}} & \multicolumn{2}{c|}{\textbf{Validation}} & \multicolumn{2}{c}{\textbf{Test}}\\
\midrule
Bar & 1,126,919 & 71.28\% & 46,115 & 77.69\%  & 49,409 & 78.92\% \\
Line  & 92,373 & 5.84\%  & 2,466 & 4.15\%  & 2,467 & 3.94\% \\
Pie  & 141,449 & 8.95\%  & 3,955 & 6.66\%  & 3,775 & 6.03\% \\
Legend  & 75,534 & 4.78\%  & 2,503 & 4.22\%  & 2,564 & 4.10\% \\
ValueAxisTitle  & 47,704 & 3.02\%  & 1,446 & 2.44\%  & 1,417 & 2.26\% \\
ChartTitle  & 69,289 & 4.38\%  & 2,127 & 3.58\%  & 2,200 & 3.51\% \\
CategoryAxisTitle  & 27,740 & 1.75\%  & 745 & 1.26\%  & 773 & 1.23\% \\
\midrule
\textbf{Total}  & 1,581,008 & & 59,357 & & 62,605 & \\
\bottomrule
\end{tabular}
}
\caption{
Category distribution in the ExcelChart400K instance segmentation dataset.
}
\label{tab:stat}
\vspace{\temp}
\end{table*}

\paragraph{Dataset}

Previous chart component recognition datasets are limited in their coverage of chart element types, making them inadequate for a comprehensive evaluation and only offering restricted chart understanding analysis for subsequent ChartQA tasks. Thus we leverage the ExcelChart400K dataset to automatically create annotations for 7 categories, including bar, line, pie, legend, value axis title, chart title, and category axis title\footnote{Category axis title refers to axis title for categorical values.}.
As the annotations in the ExcelChart400K dataset are formatted for key point detection, we convert them to an instance segmentation format.
Each object and its category remains the same, but we now create segmentation masks from the given key points.
We store segmentation masks in the format of \emph{polygons}, meaning that for each chart component, we create a collection $\mathcal{P} = [x_1, y_1, \ldots, x_n, y_n]$ defining the adjacent vertices $(x_i, y_i)$ of the associated polygon.
The procedure for converting these annotations to an instance segmentation format is outlined in Appendix~\ref{sec:annotation}. Details about the category distribution of the annotations can be found in Table~\ref{tab:stat}.

\paragraph{Evaluation Metrics}
We evaluate the performance of chart component recognition with three metrics: (1) mAP (mean Average Precision) is a common metric for evaluating object detection and instance segmentation tasks. It calculates the average precision for each class across all recall levels and then averages these values to get a single score. (2) mAP50 specifically refers to the mean Average Precision calculated at an intersection over union (IoU) threshold of 0.5. (3) mAP75 is similar to mAP50 but calculated at an IoU threshold of 0.75, and thus a stricter metric requiring a higher amount of overlap.

\subsection{Chart Question Answering}
\paragraph{Baselines}
We benchmark our model against twelve established baselines to assess its performance: \textbf{T5} \cite{raffel_exploring_2020}, a unified sequence-to-sequence Transformer model, well known for its state-of-the-art results in text-to-text tasks; \textbf{VL-T5} \cite{cho_unifying_2021}, which modifies T5 to address Vision Language tasks by generating text from multimodal inputs; \textbf{TaPas} \cite{masry_chartqa_2022} model and its extension, \textbf{VisionTapas} \cite{masry_chartqa_2022}, recognized for their efficacy in table encoding; the latter has been adapted for chart-based question answering with an additional table input; in addition, we also compare our model with the pretrained versions of \textbf{VisionTapas} and \textbf{VL-T5} on PlotQA \cite{methani_plotqa_2020} with chart VQA tasks; \textbf{ChartReader} \cite{cheng_chartreader_2023}, which is based on the T5 model and leverages extracted chart information as input; \textbf{Pix2Struct} \cite{lee_pix2struct_2022}, which facilitates pixel-to-text design for visually rich document understanding; \textbf{ChartT5} \cite{zhou_enhanced_2023}, which is pretrained based on VL-T5 with a masked chart-table pairing objective using chart VQA data; two excessively  pretrained models with additional chart comprehension data: \textbf{Matcha} \cite{liu_matcha_2022} and \textbf{UniChart} \cite{masry_unichart_2023} and two large language models (LLM): \textbf{PaLI-17B} \cite{chen_pali_2023} and \textbf{LLaVA1.5-13B} \cite{liu_improved_2024}. \textbf{Donut} \cite{kim_ocr-free_2022}, an OCR-oriented vision-encoder-decoder model, is distinguished by its proficiency in vision-based question answering and summarization. These baselines were selected for their relevance and demonstrated effectiveness in their respective domains, providing a comprehensive reference for comparing our model's capabilities.

\paragraph{Dataset}
The ChartQA dataset \cite{masry_chartqa_2022} includes three prevalent chart types -- a total of $18,337$ bar charts, $2,800$ line charts, and $808$ pie charts. The dataset consists of two subsets: a human-written question set and a machine-generated question set. Details about the distribution of data in the ChartQA dataset can be found in Table~\ref{tab:charqa} in Appendix~\ref{sec:appendix}.

\paragraph{Evaluation Metrics}
To evaluate our approach, we follow previous works \cite{masry_chartqa_2022,luo_chartocr_2021} and utilize Relaxed Accuracy (RA) for ChartQA. Relaxed Accuracy allows a minor 5\% inaccuracy of numeric answers to account for errors in the data extraction process. For non-numerical answers, Relaxed Accuracy requires an exact match (ignoring case) for the answer to be counted as correct. 
\section{Results}

\subsection{Chart Component Recognition}

Table \ref{tab:all} illustrates the performance of all the models employed in chart element detection and classification. Among them, \Model{} exhibits the most impressive performance across all mAP evaluation metrics. Notably, the results attained by \Model{} remain consistently superior across all categories, attaining either the top or second best mAP score, as shown in Table \ref{tab:cate}. Note that the line component emerges as the most challenging category due to its distinctiveness from other categories and its unique structural characteristics. 

Compared to DAT, \Model{} achieves better performance, particularly in categories that are challenging for DAT, such as lines. Experimental results indicate that the narrower the line width we annotate, the more difficult it is for DAT to predict the line segments accurately. However, \Model{} consistently predicts the line segments accurately, regardless of how narrow the line annotations are. Additionally, compared to Mask2Former, \Model{} increases the accuracy for bar segments by 10.0\%. In complex scenarios such as stacked bar charts and overlapping line charts, \Model{} more accurately detects the correct number of components. Detailed comparison examples are provided in Figures~\ref{fig:pie},~\ref{fig:bar},~and~\ref{fig:line} in Appendix~\ref{sec:appendix}.
These examples reveal the shortcomings of competing models, such as the inability to predict smooth pie edges, small bars, or steep lines. Though Mask2former performs relatively well, it occasionally misses stacked bar predictions and overcompensates on the width of the lines.

\begin{table}
\small \centering
\scalebox{0.92}{
\begin{tabular}{l|l|l|l}
\toprule
\textbf{Model} & \textbf{mAP} & \textbf{mAP50} & \textbf{mAP75}\\
\midrule
Mask R-CNN \cite{he_mask_2018} & 48.5& 62.9& 54.8 \\
SOLOv2  \cite{wang_solov2_2020} & 18.4& 37.0& 16.1 \\
DAT \cite{xia_dat_2023} & 53.5& 64.3& 59.0\\
Mask2former \cite{cheng_masked-attention_2022} & 61.7& 82.4& 65.5 \\
\midrule
\textbf{\Model{}} & \textbf{64.9}& \textbf{85.0}& \textbf{69.1}\\
\bottomrule
\end{tabular}
}
\caption{
Comparison with baselines on chart component recognition.
}
\label{tab:all}
\vspace{\temp}
\end{table}

\begin{table*}
\small \centering
\scalebox{0.92}{
\begin{tabular}{l|c|c|c|c|c|c|c|c}
\toprule
\textbf{Model} & \textbf{Bar} & \textbf{Line} & \textbf{Pie}& \textbf{Legend} & \textbf{ValueAT} &\textbf{ChartTitle}  & \textbf{CategoryAT} & \textbf{Average}\\
\midrule
Mask R-CNN \cite{he_mask_2018} & 69.6& 0.1& 61.6& 69.7& 37.2& 65.2& 36.1 & 48.5\\
SOLOv2  \cite{wang_solov2_2020} & 3.1& 0.0& 33.9& 29.1& 10.9& 42.7& 9.3 &18.4 \\
DAT \cite{xia_dat_2023} & \textbf{75.0}& 0.6& 63.3& 81.2& 40.9& \textbf{71.8}& 41.6 & 53.5\\
Mask2former \cite{cheng_masked-attention_2022} & 63.1& 28.8& 79.8& 80.3& 52.8& 69.6& \textbf{57.7} & 61.7 \\\midrule
\textbf{\Model{}} & 73.1&\textbf{36.3} &\textbf{84.0}&\textbf{81.9}&\textbf{53.1}&71.6&54.2&\textbf{64.9} \\\bottomrule
\end{tabular}
}
\caption{
Chart component recognition results (mAP) for all categories.
AT denotes AxisTitle. (Best performances are highlighted in bold.) 
} 
\label{tab:cate}
\end{table*}

\subsection{Chart Question Answering}
The results of the chart visual question answering task are summarized in Table \ref{tab:qa}. \Modelvqa{} significantly outperforms all non-pretraining baseline methods and some of the pertaining and LLM-based baselines. 
However, our model underperforms compared to extensively pre-trained models like MatCha and Unichart. It is important to note that MatCha is pre-trained on 25 million chart-related data points, while Unichart leverages 13 million high-resolution chart data entries. This extensive pre-training enables these models to learn a wider variety of patterns and nuances, leading to superior performance. In contrast, our model involves minimal pre-training on ExcelChart400K, which may not capture the same level of detail and complexity, resulting in relatively lower performance.

\begin{figure*}[ht]
    \centering
    \includegraphics[width=\textwidth]{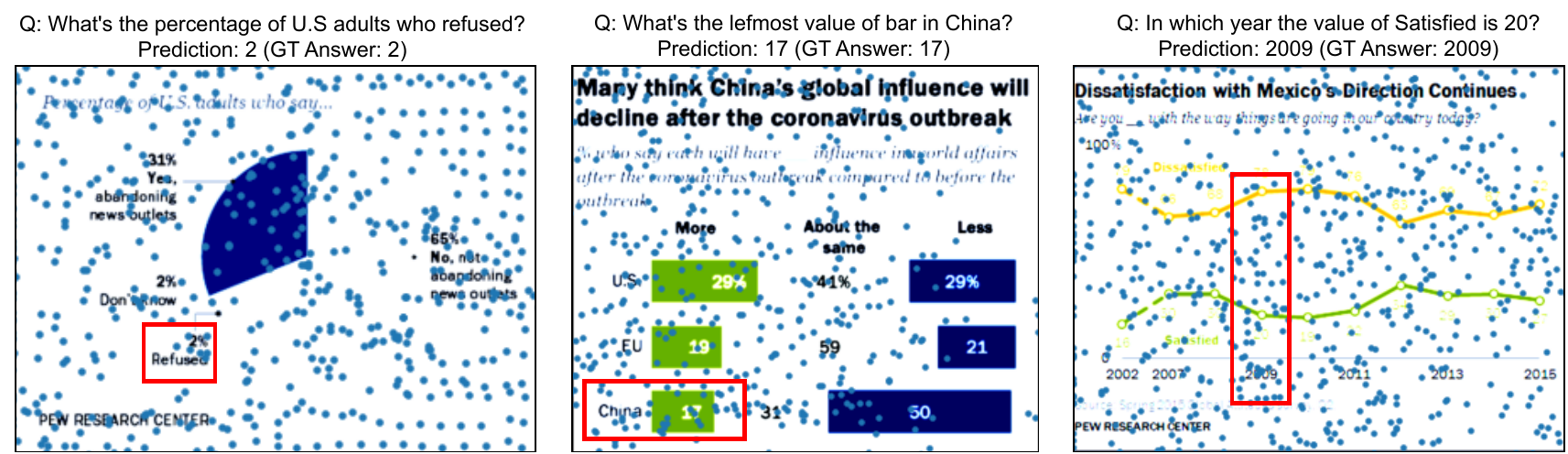}
    \caption{Qualitative examples on question-guided deformed points}
    \label{fig:qavis}
    \vspace{\temp}
\end{figure*}

We further visualize deformed points sampled by the bilinear sampling step in \Modelvqa{} in Figure ~\ref{fig:qavis} to demonstrate the effectiveness of the proposed framework. We present three human-written ChartQA examples with different types of charts (bar, line, and pie) from the test dataset. These visualizations illustrate a significant correlation between deformed points and regions containing potential answers, particularly the correct answer (highlighted by the red box in Figure ~\ref{fig:qavis}).
{A common observation is that the deformed points tend to cluster around text and chart data element regions while being sparse or uniformly spaced in the blank areas. We initialize our deformable points uniformly spaced, and they move towards nearby visual elements or stay still when nothing is around. The deformed points align with the question text and visual traits; for instance, in the line graph, points cluster around the answer ``2009'' region, locating the answer correctly; in the bar graph, where the question asks for the ``leftmost value'', most points shift to the left, demonstrating the model's ability to understand and follow the question; in the pie graph, even there are two ``2\%'' texts, the points are clustered on the one that matches the question description ``refused''. More examples can be found in Figure~\ref{fig:vis} in Appendix~\ref{sec:appendix}. This demonstrates that our proposed QDCAt enhances the model's reasoning ability through the movement of deformable points. }

\begin{table}
\small \centering
\scalebox{0.92}{
\begin{tabular}{l|c|c|c}
\toprule
\textbf{Model} & \textbf{OCR}& \textbf{Size} & \textbf{Accuracy}\\
\midrule
Donut \cite{kim_ocr-free_2022} & \xmark &201M & 41.8\\
VisionTaPas \cite{masry_chartqa_2022} & \cmark &110M & 45.5\\
TaPas \cite{herzig_tapas_2020} & \cmark &110M & 41.3\\
T5 \cite{raffel_exploring_2020} & \cmark &220M & 41.0\\
VL-T5 \cite{cho_unifying_2021} & \cmark &220M & 41.6\\
ChartReader \cite{cheng_chartreader_2023} & \cmark &220M & 52.6\\
Pix2Struct \cite{lee_pix2struct_2022} & \xmark & 282M & 56.0\\
\midrule
VL-T5$_\text{pre}$ \cite{cho_unifying_2021} & \cmark &220M & 51.8\\
VisionTaPas$_\text{pre}$ \cite{masry_chartqa_2022} & \cmark &110M & 47.1\\
ChartT5 \cite{zhou_enhanced_2023} & \cmark &220M & 53.2\\
MatCha \cite{liu_matcha_2022} & \xmark &300M & \textbf{64.2}\\
UniChart \cite{masry_unichart_2023} & \xmark &201M & \textbf{66.2}\\
\midrule
PaLI-17B \cite{chen_pali_2023} &\xmark &17B & 47.6\\
LLaVA1.5-13B \cite{liu_improved_2024} &\xmark &13B & 55.3\\
\midrule
\textbf{\Modelvqa{}} &\xmark &259M & \textbf{57.2}\\
\bottomrule
\end{tabular}
}
\caption{
Comparison with baselines on ChartQA. 
}
\label{tab:qa}
\vspace{\temp}
\end{table}

\subsection{Ablation Studies}

To demonstrate the contributions of different components to the overall performance of \Modelvqa{}, we conduct ablation studies and experiment with two additional methods to combine the features from the two image encoders.

\begin{itemize}
\item \Modelvqa{} - QON: We remove the question-guided offset network (QON) and bilinear sampling from \Modelvqa{}. 
\item \Modelvqa{}$_{\text{Concat}}$: Instead of using the deformable co-attention block, we test concatenation over the channel dimension, denoted by $\bsb{\tilde{x}} \oplus \bsb{y}$. 
\item \Modelvqa{}$_{\text{CNN}}$: We further experiment with concatenation followed by a CNN layer to adaptively fuse the spatial features, denoted by $\text{CNN}(\bsb{\tilde{x}} \oplus \bsb{y})$.
\end{itemize}

The experimental results shown in Table~\ref{tab:qaa} demonstrate that without our proposed deformable co-attention, using CNN fusion (\Modelvqa{}$_{\text{CNN}}$) decreases performance by 5.52\%. Similarly, removing the CNN layer and using simple concatenation (\Modelvqa{}$_{\text{Concat}}$) results in a further drop in accuracy to 49.92\%. Additionally, the question-guided offset network (QON) helps the model focus on visual content closely relevant to the question. When this component is removed (\Modelvqa{} - QON), accuracy decreases by 1.64\%. This setting is especially influential for human-written questions and less so for machine-generated questions. Human-written questions emphasize visual and logical reasoning and pose significant challenges for previous work. We provide baseline machine and human performance in Table~\ref{tab:qahm} in Appendix~\ref{sec:appendix}. Our model demonstrates that using the question as an early multimodal fusion cue enhances the ability to answer more complex questions.
\begin{table}[ht]
\small \centering
\scalebox{0.92}{
\begin{tabular}{l|l|l|l}
\toprule
\textbf{Model} & \textbf{Human}& \textbf{Machine} & \textbf{Average} \\
\midrule
\textbf{\Modelvqa{}}  & \textbf{34.9} &\textbf{79.4}& \textbf{57.2}\\
\Modelvqa{} - QON & 31.2&79.2&55.2 \\
\Modelvqa{}$_{\text{Concat}}$  & 26.2 &73.6  & 49.9\\
\Modelvqa{}$_{\text{CNN}}$ & 27.2 &76.0 & 51.6\\

\bottomrule
\end{tabular}
}
\caption{
Ablation results on ChartQA
}
\label{tab:qaa}
\vspace{\temp}
\end{table}

\section{Conclusion}

In this study, we address ChartQA by enhancing chart component recognition and proposing a novel question-aware attention fusion module. 
We introduce \Model{}, a unified network designed to handle multiple chart comprehension tasks across various chart types in an end-to-end manner.
    This innovative architecture is carefully crafted to handle the intricate nuances associated with diverse chart components, offering a robust solution for accurate and reliable instance segmentation. 
We further propose \Modelvqa{}, which integrates a novel Question-guided Deformable Co-Attention (QDCAt) fusion block to align question-aware chart features extracted by \Model{} with general-purpose multimodal encoder features. This approach explores new possibilities in multimodal fusion and enhances the guidance derived from the question. Through extensive experimentation and evaluation, our approach demonstrates exceptional performance, highlighting its effectiveness in addressing ChartQA challenges.


{\small
\bibliographystyle{ieee_fullname}
\bibliography{latex/custom}

\begin{thebibliography}{10}\itemsep=-1pt

\bibitem{agrawal_vqa_2016}
Aishwarya Agrawal, Jiasen Lu, Stanislaw Antol, Margaret Mitchell, C.~Lawrence Zitnick, Dhruv Batra, and Devi Parikh.
\newblock {VQA}: {Visual} {Question} {Answering}, Oct. 2016.
\newblock arXiv:1505.00468 [cs].

\bibitem{chen_pali_2023}
Xi Chen, Xiao Wang, Soravit Changpinyo, A.~J. Piergiovanni, Piotr Padlewski, Daniel Salz, Sebastian Goodman, Adam Grycner, Basil Mustafa, Lucas Beyer, Alexander Kolesnikov, Joan Puigcerver, Nan Ding, Keran Rong, Hassan Akbari, Gaurav Mishra, Linting Xue, Ashish Thapliyal, James Bradbury, Weicheng Kuo, Mojtaba Seyedhosseini, Chao Jia, Burcu~Karagol Ayan, Carlos Riquelme, Andreas Steiner, Anelia Angelova, Xiaohua Zhai, Neil Houlsby, and Radu Soricut.
\newblock {PaLI}: {A} {Jointly}-{Scaled} {Multilingual} {Language}-{Image} {Model}, June 2023.
\newblock arXiv:2209.06794 [cs].

\bibitem{cheng_masked-attention_2022}
Bowen Cheng, Ishan Misra, Alexander~G. Schwing, Alexander Kirillov, and Rohit Girdhar.
\newblock Masked-attention {Mask} {Transformer} for {Universal} {Image} {Segmentation}, June 2022.
\newblock arXiv:2112.01527 [cs].

\bibitem{cheng_chartreader_2023}
Zhi-Qi Cheng, Qi Dai, Siyao Li, Jingdong Sun, Teruko Mitamura, and Alexander~G. Hauptmann.
\newblock {ChartReader}: {A} {Unified} {Framework} for {Chart} {Derendering} and {Comprehension} without {Heuristic} {Rules}, Apr. 2023.
\newblock arXiv:2304.02173 [cs].

\bibitem{cho_unifying_2021}
Jaemin Cho, Jie Lei, Hao Tan, and Mohit Bansal.
\newblock Unifying {Vision}-and-{Language} {Tasks} via {Text} {Generation}, May 2021.
\newblock arXiv:2102.02779 [cs].

\bibitem{choi_visualizing_2019}
Jinho Choi, Sanghun Jung, Deok~Gun Park, Jaegul Choo, and Niklas Elmqvist.
\newblock Visualizing for the {Non}‐{Visual}: {Enabling} the {Visually} {Impaired} to {Use} {Visualization}.
\newblock {\em Computer Graphics Forum}, 38(3):249--260, June 2019.

\bibitem{dai_deformable_2017}
Jifeng Dai, Haozhi Qi, Yuwen Xiong, Yi Li, Guodong Zhang, Han Hu, and Yichen Wei.
\newblock Deformable {Convolutional} {Networks}, June 2017.
\newblock arXiv:1703.06211 [cs].

\bibitem{10155181}
Ning Ding, Ce Zhang, and Azim Eskandarian.
\newblock Saliendet: A saliency-based feature enhancement algorithm for object detection for autonomous driving.
\newblock {\em IEEE Transactions on Intelligent Vehicles}, 9(1):2624--2635, 2024.

\bibitem{ding_mukea_2022}
Yang Ding, Jing Yu, Bang Liu, Yue Hu, Mingxin Cui, and Qi Wu.
\newblock {MuKEA}: {Multimodal} {Knowledge} {Extraction} and {Accumulation} for {Knowledge}-based {Visual} {Question} {Answering}.
\newblock In {\em 2022 {IEEE}/{CVF} {Conference} on {Computer} {Vision} and {Pattern} {Recognition} ({CVPR})}, pages 5079--5088, New Orleans, LA, USA, June 2022. IEEE.

\bibitem{dosovitskiy_image_2021}
Alexey Dosovitskiy, Lucas Beyer, Alexander Kolesnikov, Dirk Weissenborn, Xiaohua Zhai, Thomas Unterthiner, Mostafa Dehghani, Matthias Minderer, Georg Heigold, Sylvain Gelly, Jakob Uszkoreit, and Neil Houlsby.
\newblock An {Image} is {Worth} 16x16 {Words}: {Transformers} for {Image} {Recognition} at {Scale}, June 2021.
\newblock arXiv:2010.11929 [cs].

\bibitem{hassani_neighborhood_2022}
Ali Hassani, Steven Walton, Jiachen Li, Shen Li, and Humphrey Shi.
\newblock Neighborhood {Attention} {Transformer}, Apr. 2022.

\bibitem{he_mask_2018}
Kaiming He, Georgia Gkioxari, Piotr Dollár, and Ross Girshick.
\newblock Mask {R}-{CNN}, Jan. 2018.
\newblock arXiv:1703.06870 [cs].

\bibitem{he_deep_2015}
Kaiming He, Xiangyu Zhang, Shaoqing Ren, and Jian Sun.
\newblock Deep {Residual} {Learning} for {Image} {Recognition}, Dec. 2015.
\newblock arXiv:1512.03385 [cs].

\bibitem{herzig_tapas_2020}
Jonathan Herzig, Paweł~Krzysztof Nowak, Thomas Müller, Francesco Piccinno, and Julian~Martin Eisenschlos.
\newblock {TAPAS}: {Weakly} {Supervised} {Table} {Parsing} via {Pre}-training.
\newblock In {\em Proceedings of the 58th {Annual} {Meeting} of the {Association} for {Computational} {Linguistics}}, pages 4320--4333, 2020.
\newblock arXiv:2004.02349 [cs].

\bibitem{kim_ocr-free_2022}
Geewook Kim, Teakgyu Hong, Moonbin Yim, Jeongyeon Nam, Jinyoung Park, Jinyeong Yim, Wonseok Hwang, Sangdoo Yun, Dongyoon Han, and Seunghyun Park.
\newblock {OCR}-free {Document} {Understanding} {Transformer}, Oct. 2022.
\newblock arXiv:2111.15664 [cs].

\bibitem{lal_lineformer_2023}
Jay Lal, Aditya Mitkari, Mahesh Bhosale, and David Doermann.
\newblock {LineFormer}: {Rethinking} {Line} {Chart} {Data} {Extraction} as {Instance} {Segmentation}, May 2023.
\newblock arXiv:2305.01837 [cs].

\bibitem{lee_pix2struct_2022}
Kenton Lee, Mandar Joshi, Iulia Turc, Hexiang Hu, Fangyu Liu, Julian Eisenschlos, Urvashi Khandelwal, Peter Shaw, Ming-Wei Chang, and Kristina Toutanova.
\newblock {Pix2Struct}: {Screenshot} {Parsing} as {Pretraining} for {Visual} {Language} {Understanding}, Oct. 2022.
\newblock arXiv:2210.03347 [cs].

\bibitem{liu_matcha_2022}
Fangyu Liu, Francesco Piccinno, Syrine Krichene, Chenxi Pang, Kenton Lee, Mandar Joshi, Yasemin Altun, Nigel Collier, and Julian~Martin Eisenschlos.
\newblock {MatCha}: {Enhancing} {Visual} {Language} {Pretraining} with {Math} {Reasoning} and {Chart} {Derendering}, Dec. 2022.
\newblock arXiv:2212.09662 [cs].

\bibitem{liu_improved_2024}
Haotian Liu, Chunyuan Li, Yuheng Li, and Yong~Jae Lee.
\newblock Improved {Baselines} with {Visual} {Instruction} {Tuning}, May 2024.
\newblock arXiv:2310.03744 [cs].

\bibitem{liu2019dataextractionchartssingle}
Xiaoyi Liu, Diego Klabjan, and Patrick NBless.
\newblock Data extraction from charts via single deep neural network, 2019.

\bibitem{liu_multilingual_2020}
Yinhan Liu, Jiatao Gu, Naman Goyal, Xian Li, Sergey Edunov, Marjan Ghazvininejad, Mike Lewis, and Luke Zettlemoyer.
\newblock Multilingual {Denoising} {Pre}-training for {Neural} {Machine} {Translation}, Jan. 2020.
\newblock arXiv:2001.08210 [cs].

\bibitem{liu_swin_2021}
Ze Liu, Yutong Lin, Yue Cao, Han Hu, Yixuan Wei, Zheng Zhang, Stephen Lin, and Baining Guo.
\newblock Swin {Transformer}: {Hierarchical} {Vision} {Transformer} using {Shifted} {Windows}, Aug. 2021.
\newblock arXiv:2103.14030 [cs].

\bibitem{luo_chartocr_2021}
Junyu Luo, Zekun Li, Jinpeng Wang, and Chin-Yew Lin.
\newblock {ChartOCR}: {Data} {Extraction} from {Charts} {Images} via a {Deep} {Hybrid} {Framework}.
\newblock In {\em 2021 {IEEE} {Winter} {Conference} on {Applications} of {Computer} {Vision} ({WACV})}, pages 1916--1924, Waikoloa, HI, USA, Jan. 2021. IEEE.

\bibitem{ma_towards_2021}
Weihong Ma, Hesuo Zhang, Shuang Yan, Guangshun Yao, Yichao Huang, Hui Li, Yaqiang Wu, and Lianwen Jin.
\newblock Towards an efficient framework for {Data} {Extraction} from {Chart} {Images}, May 2021.
\newblock arXiv:2105.02039 [cs].

\bibitem{masry_chartqa_2022}
Ahmed Masry, Xuan~Long Do, Jia~Qing Tan, Shafiq Joty, and Enamul Hoque.
\newblock {ChartQA}: {A} {Benchmark} for {Question} {Answering} about {Charts} with {Visual} and {Logical} {Reasoning}.
\newblock In {\em Findings of the {Association} for {Computational} {Linguistics}: {ACL} 2022}, pages 2263--2279, Dublin, Ireland, May 2022. Association for Computational Linguistics.

\bibitem{masry_unichart_2023}
Ahmed Masry, Parsa Kavehzadeh, Xuan~Long Do, Enamul Hoque, and Shafiq Joty.
\newblock {UniChart}: {A} {Universal} {Vision}-language {Pretrained} {Model} for {Chart} {Comprehension} and {Reasoning}, Oct. 2023.
\newblock arXiv:2305.14761 [cs].

\bibitem{methani_plotqa_2020}
Nitesh Methani, Pritha Ganguly, Mitesh~M. Khapra, and Pratyush Kumar.
\newblock {PlotQA}: {Reasoning} over {Scientific} {Plots}, Feb. 2020.
\newblock arXiv:1909.00997 [cs].

\bibitem{p_lineex_2023}
Shivasankaran~V P, Muhammad Yusuf~Hassan, and Mayank Singh.
\newblock {LineEX}: {Data} {Extraction} from {Scientific} {Line} {Charts}.
\newblock In {\em 2023 {IEEE}/{CVF} {Winter} {Conference} on {Applications} of {Computer} {Vision} ({WACV})}, pages 6202--6210, Waikoloa, HI, USA, Jan. 2023. IEEE.

\bibitem{qi2023art}
Jingyuan Qi, Zhiyang Xu, Ying Shen, Minqian Liu, Di Jin, Qifan Wang, and Lifu Huang.
\newblock The art of socratic questioning: Recursive thinking with large language models.
\newblock {\em arXiv preprint arXiv:2305.14999}, 2023.

\bibitem{raffel_exploring_2020}
Colin Raffel, Noam Shazeer, Adam Roberts, Katherine Lee, Sharan Narang, Michael Matena, Yanqi Zhou, Wei Li, and Peter~J. Liu.
\newblock Exploring the {Limits} of {Transfer} {Learning} with a {Unified} {Text}-to-{Text} {Transformer}, July 2020.
\newblock arXiv:1910.10683 [cs, stat].

\bibitem{reddy2022mumuqa}
Revant~Gangi Reddy, Xilin Rui, Manling Li, Xudong Lin, Haoyang Wen, Jaemin Cho, Lifu Huang, Mohit Bansal, Avirup Sil, Shih-Fu Chang, et~al.
\newblock Mumuqa: Multimedia multi-hop news question answering via cross-media knowledge extraction and grounding.
\newblock In {\em Proceedings of the AAAI Conference on Artificial Intelligence}, volume~36, pages 11200--11208, 2022.

\bibitem{ren_faster_2016}
Shaoqing Ren, Kaiming He, Ross Girshick, and Jian Sun.
\newblock Faster {R}-{CNN}: {Towards} {Real}-{Time} {Object} {Detection} with {Region} {Proposal} {Networks}, Jan. 2016.
\newblock arXiv:1506.01497 [cs].

\bibitem{schmidhuber_deep_2015}
Juergen Schmidhuber.
\newblock Deep {Learning} in {Neural} {Networks}: {An} {Overview}.
\newblock {\em Neural Networks}, 61:85--117, Jan. 2015.
\newblock arXiv:1404.7828 [cs].

\bibitem{shen2024multimodal}
Ying Shen, Zhiyang Xu, Qifan Wang, Yu Cheng, Wenpeng Yin, and Lifu Huang.
\newblock Multimodal instruction tuning with conditional mixture of lora.
\newblock {\em arXiv preprint arXiv:2402.15896}, 2024.

\bibitem{wang_solov2_2020}
Xinlong Wang, Rufeng Zhang, Tao Kong, Lei Li, and Chunhua Shen.
\newblock {SOLOv2}: {Dynamic} and {Fast} {Instance} {Segmentation}, Oct. 2020.
\newblock arXiv:2003.10152 [cs].

\bibitem{xia_dat_2023}
Zhuofan Xia, Xuran Pan, Shiji Song, Li~Erran Li, and Gao Huang.
\newblock {DAT}++: {Spatially} {Dynamic} {Vision} {Transformer} with {Deformable} {Attention}, Sept. 2023.
\newblock arXiv:2309.01430 [cs].

\bibitem{xu2024vision}
Zhiyang Xu, Chao Feng, Rulin Shao, Trevor Ashby, Ying Shen, Di Jin, Yu Cheng, Qifan Wang, and Lifu Huang.
\newblock Vision-flan: Scaling human-labeled tasks in visual instruction tuning.
\newblock {\em arXiv preprint arXiv:2402.11690}, 2024.

\bibitem{xu2022multiinstruct}
Zhiyang Xu, Ying Shen, and Lifu Huang.
\newblock Multiinstruct: Improving multi-modal zero-shot learning via instruction tuning.
\newblock {\em arXiv preprint arXiv:2212.10773}, 2022.

\bibitem{xue_chartdetr_2023}
Wenyuan Xue, Dapeng Chen, Baosheng Yu, Yifei Chen, Sai Zhou, and Wei Peng.
\newblock {ChartDETR}: {A} {Multi}-shape {Detection} {Network} for {Visual} {Chart} {Recognition}, Aug. 2023.
\newblock arXiv:2308.07743 [cs].

\bibitem{yang_stacked_2016}
Zichao Yang, Xiaodong He, Jianfeng Gao, Li Deng, and Alex Smola.
\newblock Stacked {Attention} {Networks} for {Image} {Question} {Answering}.
\newblock In {\em 2016 {IEEE} {Conference} on {Computer} {Vision} and {Pattern} {Recognition} ({CVPR})}, pages 21--29, Las Vegas, NV, USA, June 2016. IEEE.

\bibitem{10.1115/IMECE2021-69975}
Ce Zhang and Azim Eskandarian.
\newblock A comparative analysis of object detection algorithms in naturalistic driving videos.
\newblock {\em ASME International Mechanical Engineering Congress and Exposition}, Volume 7B: Dynamics, Vibration, and Control:V07BT07A018, 11 2021.

\bibitem{9564709}
Ce Zhang, Azim Eskandarian, and Xuelai Du.
\newblock Attention-based neural network for driving environment complexity perception.
\newblock In {\em 2021 IEEE International Intelligent Transportation Systems Conference (ITSC)}, pages 2781--2787, 2021.

\bibitem{Zhang_2023_CVPR}
Ce Zhang, Chengjie Zhang, Yiluan Guo, Lingji Chen, and Michael Happold.
\newblock Motiontrack: End-to-end transformer-based multi-object tracking with lidar-camera fusion.
\newblock In {\em Proceedings of the IEEE/CVF Conference on Computer Vision and Pattern Recognition (CVPR) Workshops}, pages 151--160, June 2023.

\bibitem{10.1007/978-3-031-25066-8_41}
Yangheng Zhao, Jun Wang, Xiaolong Li, Yue Hu, Ce Zhang, Yanfeng Wang, and Siheng Chen.
\newblock Number-adaptive prototype learning for 3d point cloud semantic segmentation.
\newblock In Leonid Karlinsky, Tomer Michaeli, and Ko Nishino, editors, {\em Computer Vision -- ECCV 2022 Workshops}, pages 695--703, Cham, 2023. Springer Nature Switzerland.

\bibitem{zhou_enhanced_2023}
Mingyang Zhou, Yi Fung, Long Chen, Christopher Thomas, Heng Ji, and Shih-Fu Chang.
\newblock Enhanced {Chart} {Understanding} via {Visual} {Language} {Pre}-training on {Plot} {Table} {Pairs}.
\newblock In {\em Findings of the {Association} for {Computational} {Linguistics}: {ACL} 2023}, pages 1314--1326, Toronto, Canada, July 2023. Association for Computational Linguistics.

\end{thebibliography}
}

\appendix

\newpage
\section{Segmentation Annotation}\label{sec:annotation}

\paragraph{Pies:}
The coordinates for three vertices of the slices are already provided as keypoints.
It remains to approximate the arc of the circle via insertion of intermediate points, roughly $5$ per radian.
To achieve this, we compute the angles and radii of the two edge vertices with respect to the center, then linearly interpolate these quantities for the intermediate points.

\paragraph{Lines:}
As the provided keypoints are placed at line centers, no information regarding the thickness of a particular line is available.
To address this while accounting for differing image sizes, we set annotation line thickness to $1\%$ of the image height.
While this parameter choice produces fairly accurate annotations in most cases, it does not necessarily correspond to the ground-truth line thickness shown in images. 
The edges of the created polygon are defined by line segments parallel to the ones in the provided line, placed at the line width distance apart from each other.
Their endpoints (which are polygon vertices) are the intersection points of two adjacent pairs of lines.
To account for acute bends in the line producing elongated 'spikes' in its segmentation mask, the line vertices at such bends are duplicated and shifted a minute distance apart.
\paragraph{Bars and others:}
Remaining chart components have rectangular annotations, for which the four vertices can be computed directly from the given parameters.

\section{Supplementary materials}\label{sec:appendix}

\begin{table}[ht]
    \centering
    \small
    \begin{tabular}{lcccc}
        \toprule
        & \multicolumn{2}{c}{Human} & \multicolumn{2}{c}{Machine} \\
        \cmidrule(lr){2-3} \cmidrule(lr){4-5}
        Split & Charts & Questions & Charts & Questions \\
        \midrule
        Training & 3,699 & 7,398 & 15,474 & 20,901 \\
        Validation & 480 & 960 & 680 & 960 \\
        Test & 625 & 1,250 & 987 & 1,250 \\
        \midrule
        Total & 4,804 & 9,608 & 17,141 & 23,111 \\
        \bottomrule
    \end{tabular}
    \caption{Distribution of data in the ChartQA dataset.}
    \label{tab:charqa}
\end{table}

\begin{table}
\small \centering
\scalebox{0.92}{
\begin{tabular}{l|c|c|c}
\toprule
\textbf{Model} & \textbf{Human}& \textbf{Machine} & \textbf{Average}\\
\midrule
Donut \cite{kim_ocr-free_2022} & - & - & 41.80\\
VisionTaPas \cite{masry_chartqa_2022} & 29.60 & 61.44 & 45.52\\
TaPas \cite{herzig_tapas_2020} & 28.72 & 53.84 & 41.28\\
T5 \cite{raffel_exploring_2020} & 25.12 & 56.96 & 41.04\\
VL-T5 \cite{cho_unifying_2021} & 26.24 & 56.88 & 41.56\\
ChartReader \cite{cheng_chartreader_2023} & - & - & 52.60\\
Pix2Struct \cite{lee_pix2struct_2022} & 30.50 & 81.60 & 56.00\\
\midrule
VL-T5$_\text{pre}$ \cite{cho_unifying_2021} & 40.08 & 63.60 & 51.84\\
VisionTaPas$_\text{pre}$ \cite{masry_chartqa_2022} & 32.52 & 61.60 & 47.08\\
ChartT5 \cite{zhou_enhanced_2023} & 31.80 & 74.40 & 53.16\\
\midrule
PaLI-17B \cite{chen_pali_2023} &30.40 &64.90 & 47.60\\
LLaVA1.5-13B \cite{liu_improved_2024} & 37.68 & 72.96 & 55.32\\
\midrule
\textbf{\Modelvqa{}} & 34.88 & 79.44 & \textbf{57.16}\\
\bottomrule
\end{tabular}
}
\vspace{-2mm}
\caption{
Comparison with baseline results on ChartQA.
}
\label{tab:qahm}
\end{table}

\begin{figure*}[ht]
    \centering
    \includegraphics[width=\textwidth]{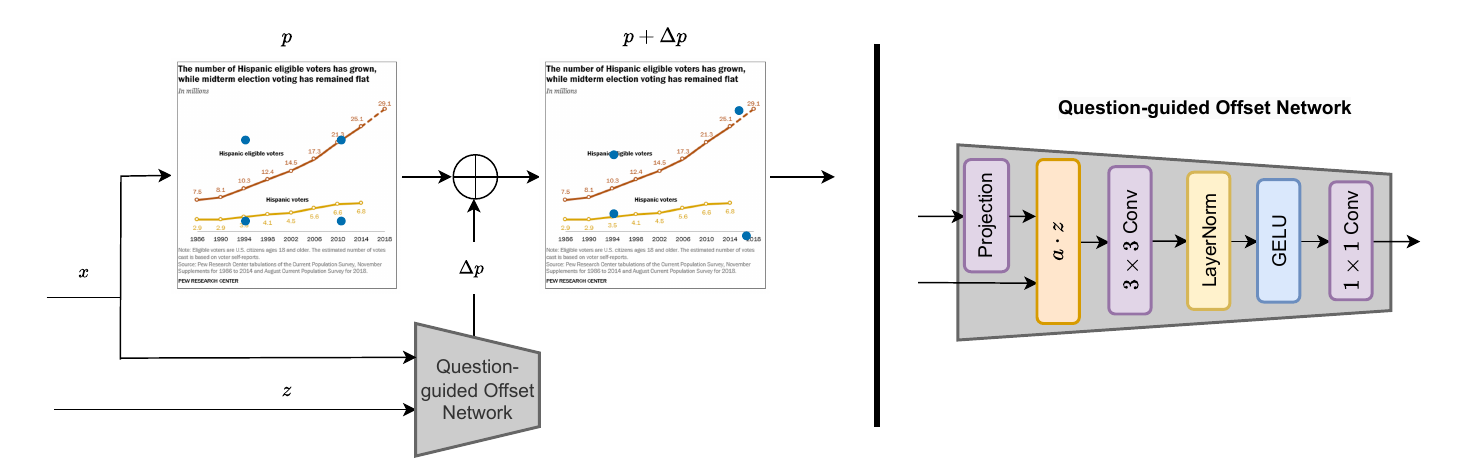}
    \caption{Question-guided Offset Network Flowchart}
    \label{fig:qgoff}
\end{figure*}

\begin{figure*}[ht]
    \centering
    \includegraphics[width=0.75\textwidth]{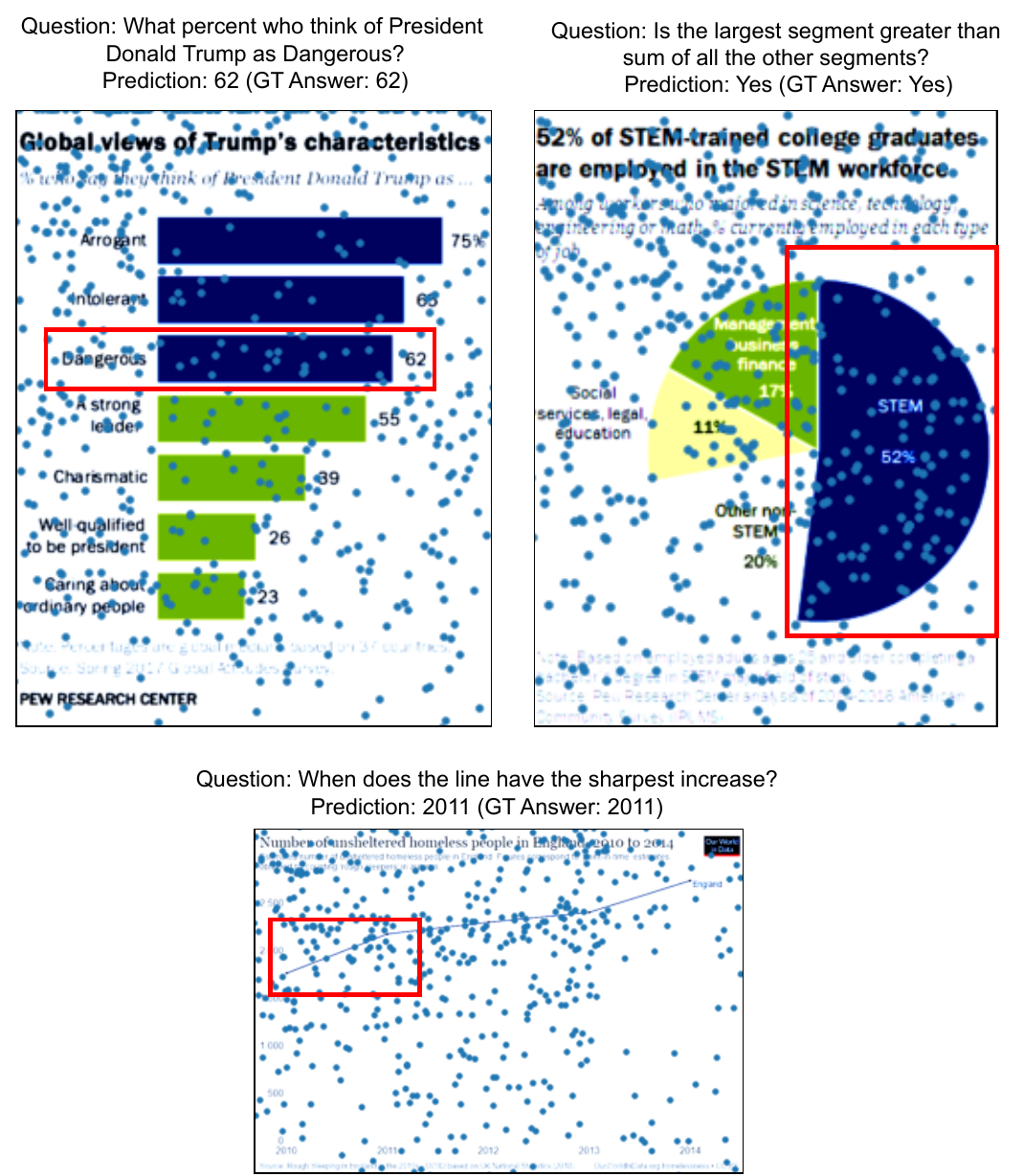}
    \caption{Visualization on question-guided deformed points}
    \label{fig:vis}
\end{figure*}

\begin{figure*}[ht]
    \centering
    \includegraphics[width=\textwidth]{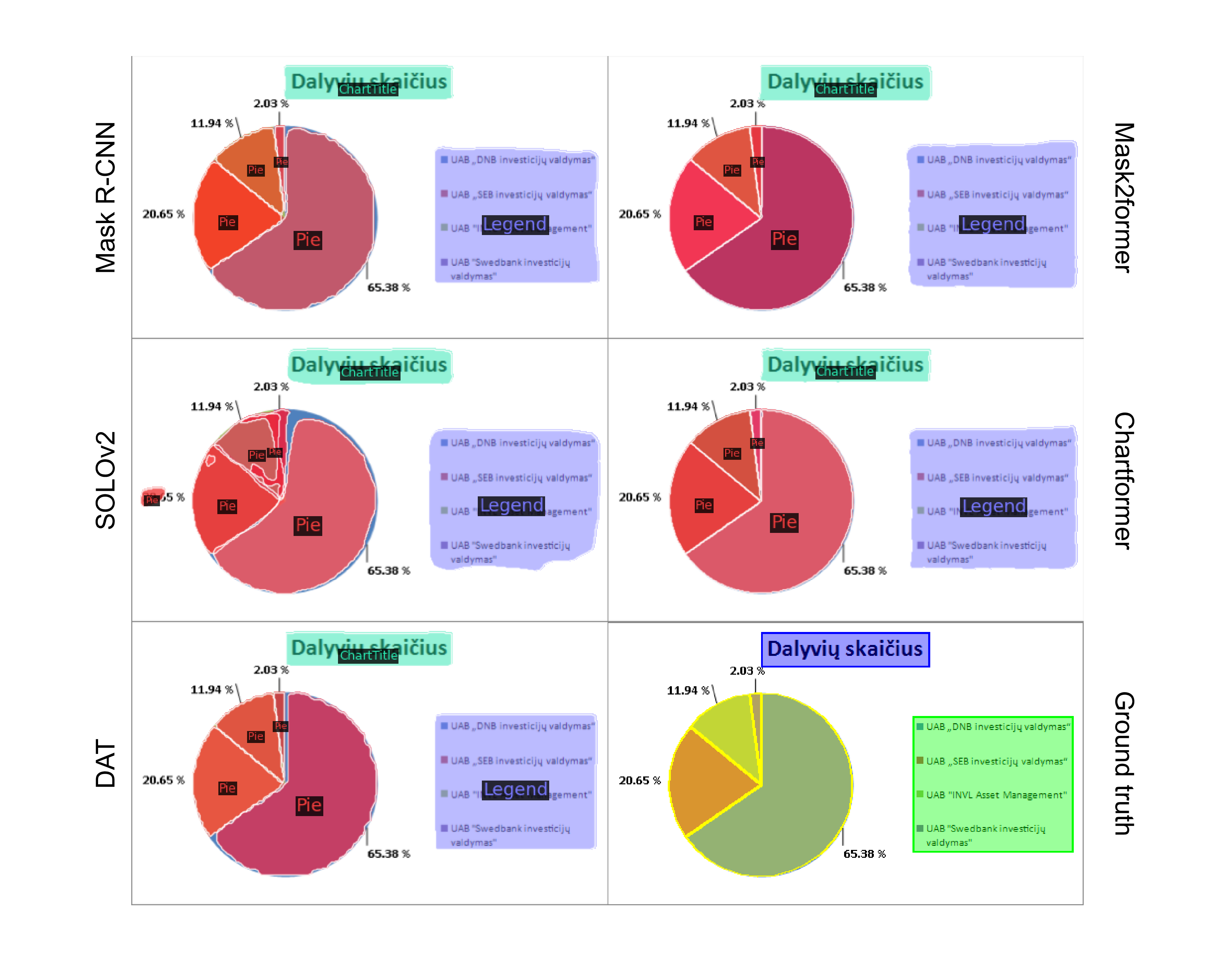}
    \caption{Pie performance case study}
    \label{fig:pie}
\end{figure*}

\begin{figure*}[ht]
    \centering
    \includegraphics[width=\textwidth]{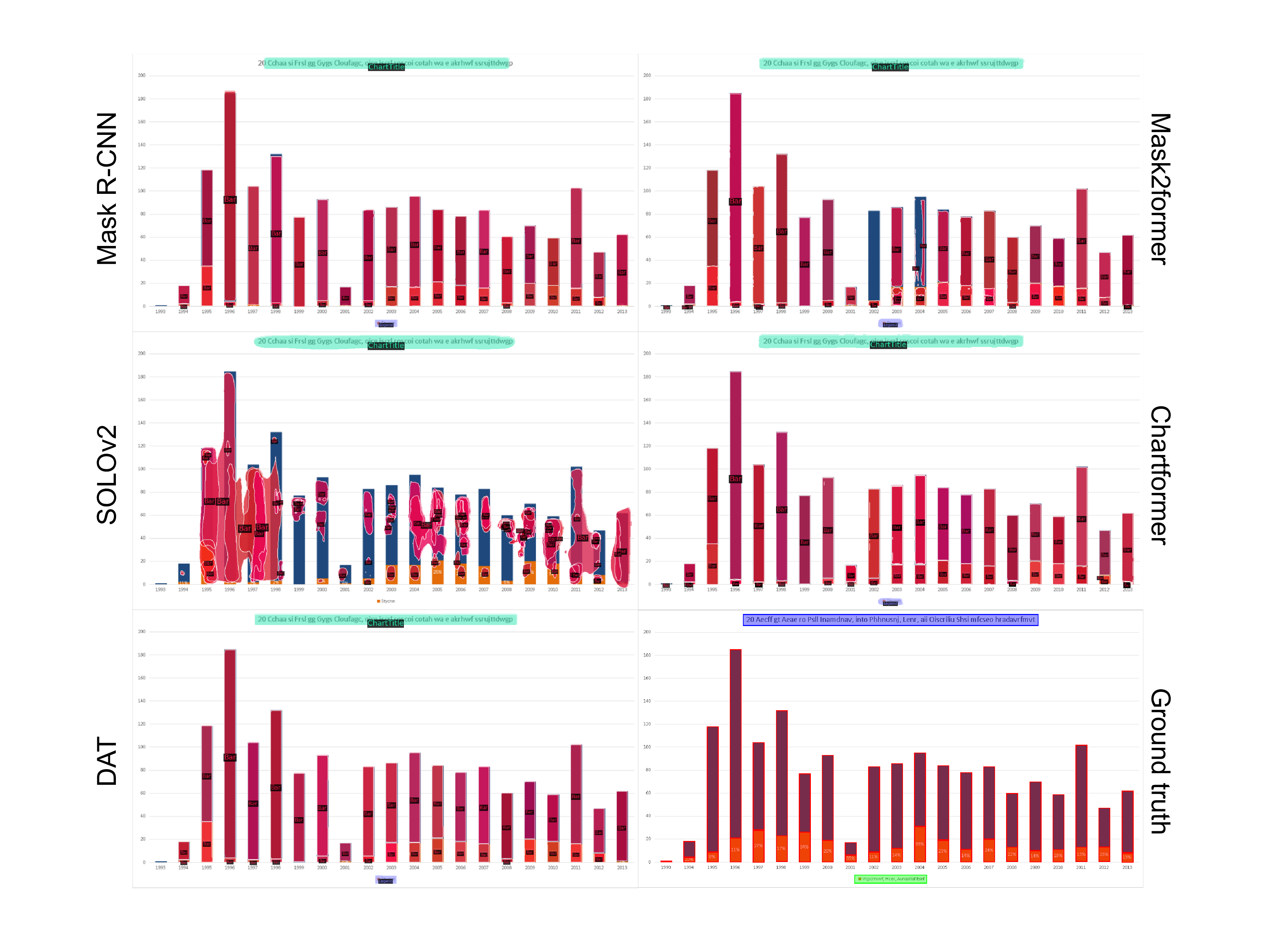}
    \caption{Bar performance case study}
    \label{fig:bar}
\end{figure*}

\begin{figure*}[ht]
    \centering
    \includegraphics[width=\textwidth]{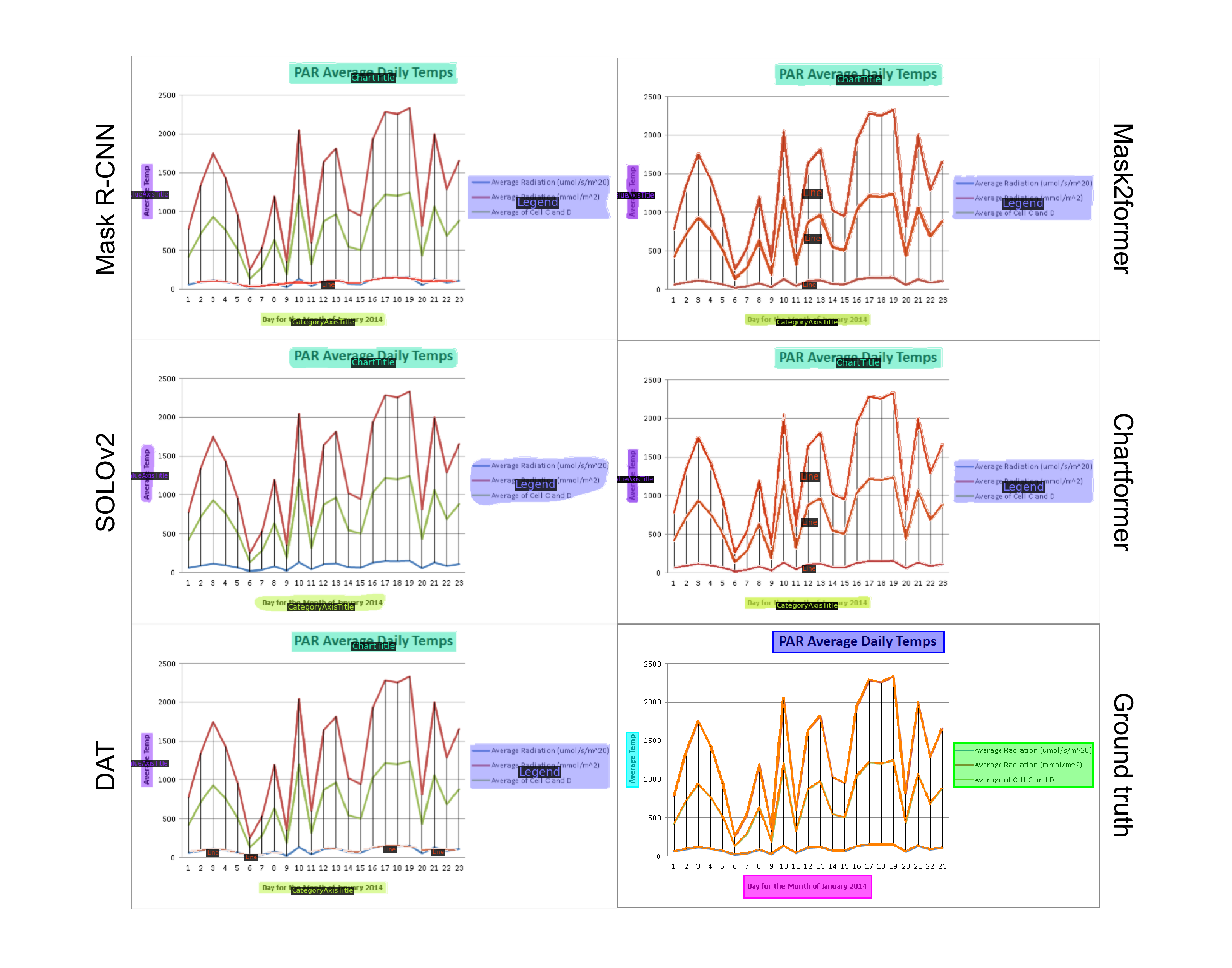}
    \caption{Line performance case study}
    \label{fig:line}
\end{figure*}
\end{document}